*Hyun-Ki JUNG* [1*]
[1] University of Seoul, Seoul, South Korea, stillhk3@uos.ac.kr
[*] Corresponding author: stillhk3@uos.ac.kr

# A text-guided vision model for enhanced recognition of small instances

**Abstract**

*As drone-based object detection technology continues to evolve, the demand is shifting from simply detecting objects to enabling users to accurately identify specific targets. For example, users can enter specific targets as prompts to accurately detect the desired objects. To address this need, an efficient text-guided object recognition model has been developed to improve the recognition of small objects. Specifically, an improved version of the existing YOLO-World model is presented. The proposed method replaces the C2f layer in the YOLOv8 backbone with a C3k2 layer, allowing for a more accurate representation of local features, especially for small objects or those with well-defined boundaries. In addition, the proposed architecture improves processing speed and efficiency by optimizing parallel processing, while contributing to a more lightweight model design. Comparative experiments on the VisDrone dataset show that the proposed model outperforms the original YOLO-World model, with precision increasing from 40.6% to 41.6%, recall from 30.8% to 31%, F1 score from 35% to 35.5%, and mAP@0.5 from 30.4% to 30.7%, confirming its improved accuracy. In addition, the model exhibits superior lightweight performance, with the number of parameters reduced from 4 million to 3.8 million and the FLOPs reduced from 15.7 billion to 15.2 billion. These results indicate that the proposed approach provides a practical and effective solution for accurate object detection in drone-based applications.*

## 1. INTRODUCTION

The drone and unmanned aerial vehicle (UAV) industry has created new opportunities across multiple sectors in recent years. With its rapid growth, the drone market is expected to account for nearly 10% of the global market. Among the various applications of drones, drone delivery is particularly highlighted as an area with high potential for future growth (Shah et al., 2024; Colpaert et al., 2022).

The rapid development of drone-based object detection technologies, combined with artificial intelligence, is further accelerating innovation. These technologies serve as key drivers for industrial progress and are being actively explored in various fields using different approaches. For example, Zhang et al. (2024) proposed an approach that integrates an evolutionary reinforcement learning agent into a fine-grained object recognition framework to optimize scale. Yang et al. (2024) explored methods to improve recognition of small objects in large scenes by incorporating object-oriented information. Xu et al. (2024) introduced a technique based on multi-scale feature fusion, while Vuong et al. (2025) conducted a comprehensive review and empirical study on drone-based wildlife detection.

Abu-Khadrah et al. (2025) proposed a novel object detection technique (ODT) that combines the whale optimization algorithm with deep reinforcement learning. Song et al. (2025) introduced an advanced drone-based IoT fusion solution for real-time safety monitoring at construction sites, providing a more effective and efficient approach to safety management. Yuan et al. (2025) developed a transformer-based multiple object tracking method (TLSH-MOT) tailored for drone-based remote sensing environments. Niu et al. (2025) developed a feature extraction network called VCBNet and a multi-attribute information integration module called VDE. To address the challenges of low accuracy and information loss in small object detection, Tao et al. (2025) proposed the MIS-YOLOv8 algorithm, while Jung et al. (2025) introduced the GhostHead network, which improves the head module of the YOLOv11 algorithm.

Looking ahead, object detection technologies using drone imagery and video need to evolve beyond basic detection by adopting a multimodal approach that integrates natural language processing (NLP) techniques.

This integration will enable more accurate and faster identification of user-specified targets, even in complex and dynamic environments.

The main contributions of this thesis are:

1. A text-guided object detection model capable of efficiently detecting small objects is presented. To achieve this, a text-guided object detection model optimized for small object detection was developed using the VisDrone dataset, which contains images captured by drones in various environments.
2. In the experiments conducted in this study, a new backbone network was introduced by replacing the C2f layers in the original YOLO-World back bone with C3k2 layers. As a result, all evaluation metrics, including precision, recall, F1 score, and mAP@0.5, showed improvements. In particular, the final evaluation metric, mAP@0.5, increased from 30.4 to 30.7, confirming an improvement in detection accuracy. In addition, the number of parameters and FLOPs also improved, indicating that the proposed model achieved better performance in terms of both efficiency and lightweight design.

## 2. RELATED WORKS

### 2.1. Object detection model

Object detection models are generally divided into two types, one-step detectors and two-step detectors. A two-stage detector performs object detection in two separate steps. In the first step, it identifies regions where objects are likely to be present. In the second step, it analyses these regions in detail to determine the exact location and class of the objects. Techniques such as selective search and sliding window are commonly used to propose candidate regions. Selective search groups similar pixels to identify potential object areas, while sliding window scans the entire image using a fixed-size rectangular window to extract object candidates. Compared to one-stage detectors, two-stage detectors involve a more complex detection process. Well-known examples of two-stage detectors include R-CNN, Fast R-CNN, Faster R-CNN, and Mask R-CNN (Glenn-Jocher, 2023; C.-Y. Wang et al., 2025; A. Wang et al., 2024; Girshick, 2015). These models have been widely used in computer vision.

In contrast, one-stage detectors perform region proposal and classification simultaneously using a convolutional neural network. These detectors perform all tasks in parallel within the convolutional layers responsible for feature extraction. As a result, they offer significantly faster detection speeds than two-stage detectors. Their simpler training and inference processes have also contributed to their widespread adoption in recent years. Representative models in this category include the You only look once (YOLO) series (Girshick, 2015; Ren, 2016; He et al., 2017; Redmon et al., 2016; Redmon & Farhadi, 2017; 2018; Bochkovskiy et al.,2020; Jocher, 2022; Li et al., 2022; Wang et al., 2023; Liu et al., 2016), single shot multibox detector (SSD), Focal Loss, and RefineDet (Lin et al., 2017; Zhang et al., 2018; Wang et al., 2020).

In this study, the feature extraction process uses only the backbone network of the YOLO model, which belongs to the category of single-shot detectors. Specifically, the proposed method improves the version eight YOLO backbone, which was originally used in the YOLO-world model to improve performance.

### 2.2. Text-guided object detection model

Deep learning and related technologies have advanced rapidly in the fields of computer vision (CV) and natural language processing (NLP). The goal of computer vision is to develop models that can extract meaningful information from visual data, such as images and videos, that can be perceived by the human eye. Natural language processing aims to create models that can understand and use human language. Accordingly, this study focuses on a multimodal model that combines these two areas, which is a text-based object detector designed to provide accurate answers when users ask questions (Jocher, 2024).

In recent years, several related studies have been conducted in this area. Wei et al. (2024) introduced an image-to-text alignment loss to remove category constraints (Shen et al., 2023). Huang et al. (2021) proposed a method for segmenting the object instance referred to by a given query sentence within a three-dimensional scene. Yi et al. integrated degradation processing of infrared and visible images with flexible interactive fusion results using a text semantic encoder and a semantic interaction fusion decoder. Chen et al. (2024) presented a new learning framework that guides the training of an image encoder using Jigsaw-based fake out-of-distribution data and rich semantic embeddings extracted from ChatGPT-generated in-distribution descriptions. Hasan et al. (2024) proposed a method for fusing visual and textual features through a context-aware attention mechanism. Liang et al. (2024) proposed an automatic data engine (AIDE) that leverages the latest advances in vision language and large language models to automatically identify problems, efficiently curate data, improve the model through automatic labelling, and validate it by generating different scenarios.

The YOLO-World model, used as a baseline in this study, is an innovative approach that provides YOLO with open vocabulary object detection capabilities through vision language modelling and large dataset pre-training (Liang et al, 2024). This model combines the efficient detection performance of the YOLOv8 backbone with the powerful text understanding and cross-modal reasoning capabilities of the CLIP model. Improvements to increase detection accuracy are introduced by using the cross-modal learning mechanism of the YOLO-World model based on CLIP (Cheng et al., 2024).

## 3. METHODOLOGY

### 3.1. The proposed YOLO-world model

The basic architecture of the proposed YOLO-World model, which includes a modified backbone network, is shown in detail in Figure 1. The input image shown in Figure 1 is an actual example from the VisDrone dataset used in the experiment (Radford et al., 2021). To describe the overall architecture, the proposed YOLO-World model first receives input texts such as "pedestrian" and "truck" from the user. The text encoder transforms the input text into embeddings, while the image encoder, based on a modified YOLOv8 backbone highlighted in red in the figure, encodes the input image into multi-scale image features. Then, the RepVL-PAN performs multi-level cross-modal fusion on both the image and text features. Finally, the proposed YOLO-World model predicts regressed bounding boxes and object embeddings corresponding to the nouns or descriptions provided in the input text.

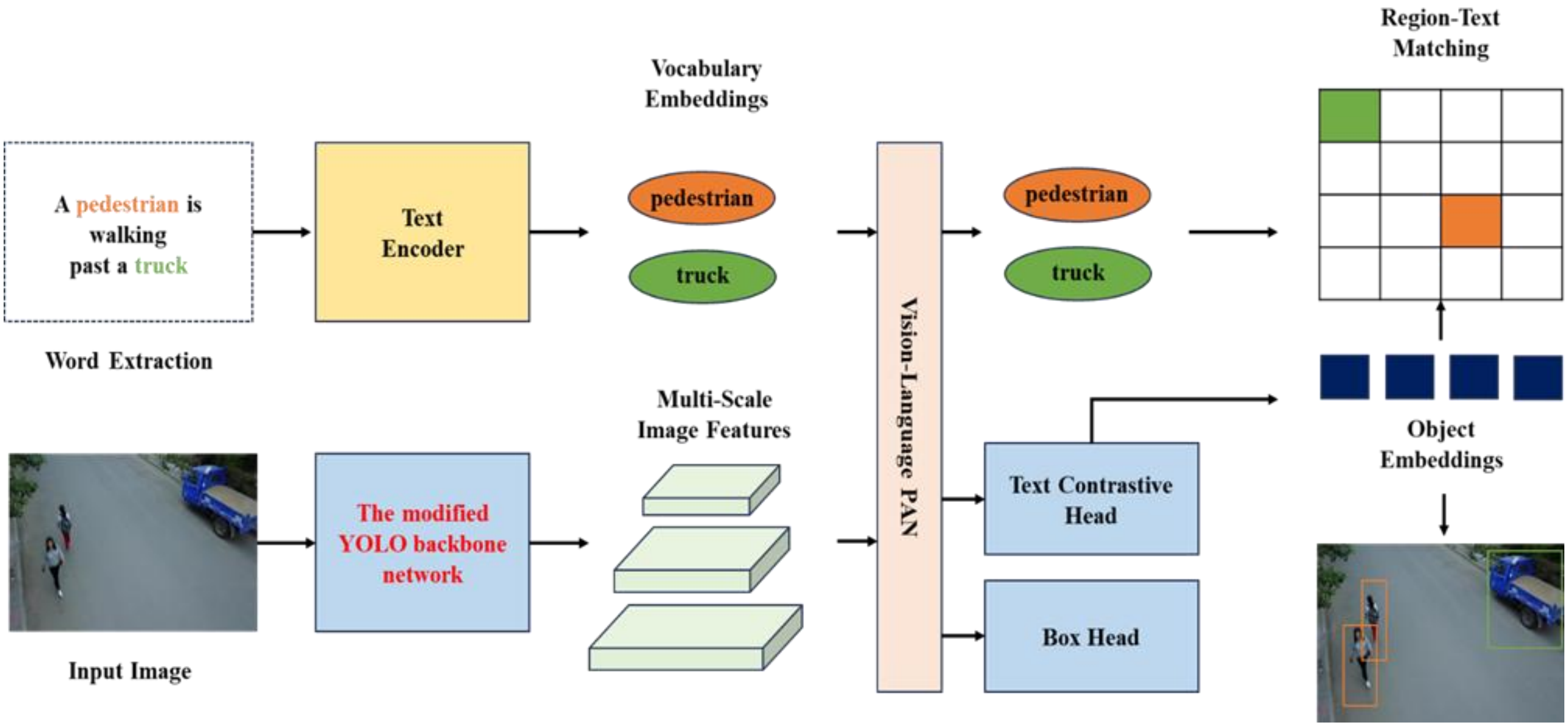

**Fig. 1. Architecture of the Proposed YOLO-World Model Used in the Experiment**

### 3.2. Improved YOLO backbone network

An improved backbone network is introduced compared to that of the original YOLO-World model, which primarily used C2f layers commonly used in YOLOv8. In this study, these layers are replaced with C3k2 layers. The C3k2 layer uses smaller 3×3 kernels to improve computational efficiency while preserving the network's ability to extract meaningful image features. First introduced in YOLOv11, the C3k2 layer is an advanced version of the Cross Stage Partial (CSP) bottleneck structure found in earlier models (AISkyEye, n.d.).

This layer optimizes information flow by splitting the feature map into a sequence of small 3×3 convolutions. These smaller kernels offer advantages such as faster processing speed and lower computational cost compared to larger kernels. By running the split feature maps through multiple convolutions and then merging them, the C3k2 layer achieves improved feature representation with fewer parameters than the C2f blocks used in YOLOv8.

The C3k layer, which is structurally similar to the C2f layer, does not involve feature map splitting. Instead, the input passes through an initial convolution block, followed by a series of n bottleneck layers with concatenations, and concludes with a final convolution block. The C3k2 layer processes information through these C3k blocks. The structure includes convolution blocks at both the beginning and the end, with several C3k blocks in between.

Finally, the outputs of the last C3k block and the previous convolution block are concatenated and run through a final convolution. This design aims to balance speed and accuracy by taking advantage of the CSP architecture. A notable strength of the C3k2 module is its enhanced ability to preserve fine-grained spatial detail, which is essential for small object detection. Its architecture uses multiple sequential 3×3 convolutional layers, enabling efficient feature extraction with minimal information loss while preserving critical edge and texture information. This capability is particularly important in small object scenarios where each pixel carries significant semantic value. In addition, the deeper path within the C3k2 module increases the number of nonlinear transformations, resulting in richer and more expressive feature representations. These architectural advantages are particularly evident in drone imagery, where objects are typically small and often partially occluded. Figure 2(a) shows the flowchart of the C3k2 layer, while Figure 2(b) shows the architecture of the proposed YOLO backbone network with the C3k2 layer applied. The modified sections are highlighted in red.

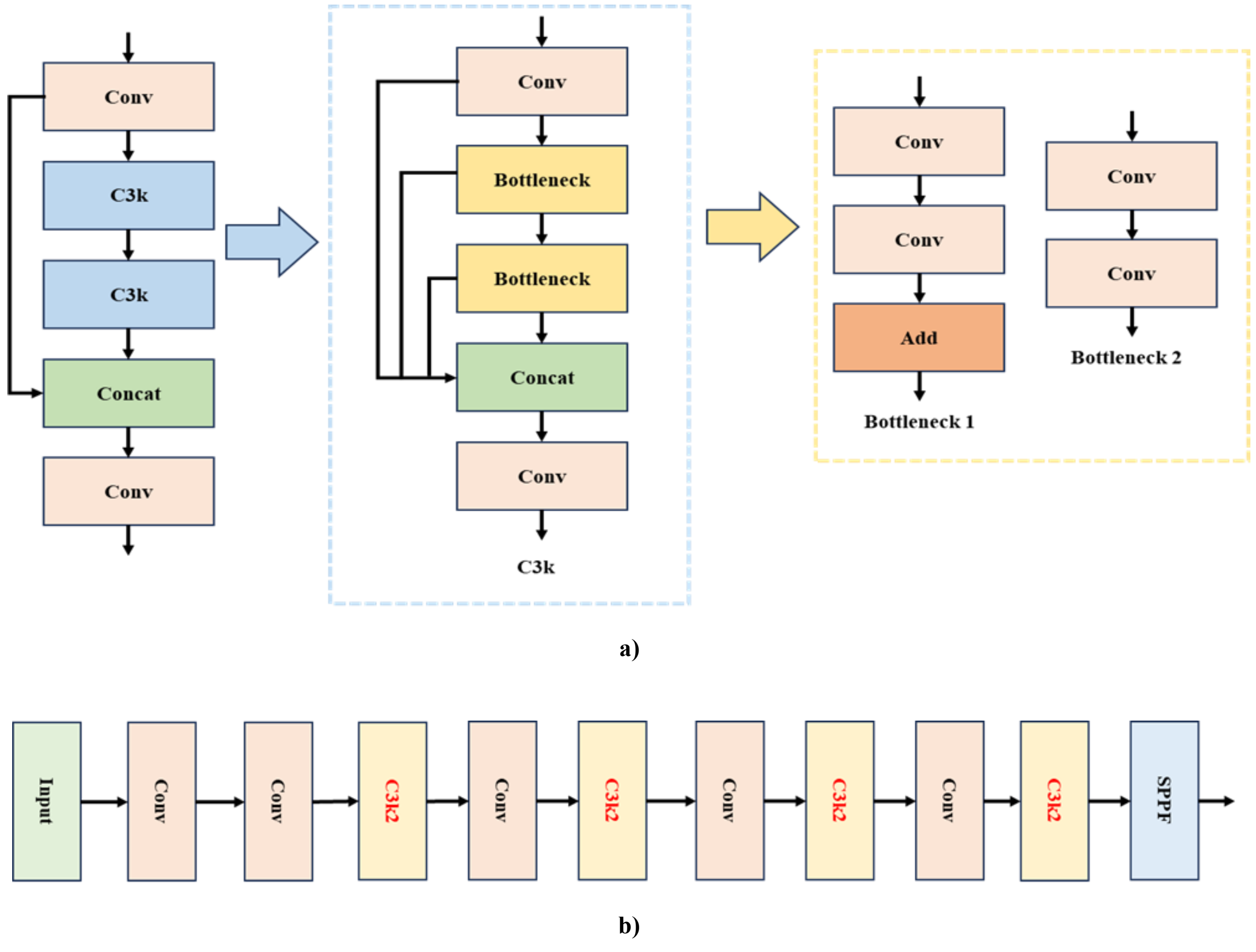

**Fig. 2. Architecture of the Backbone Network in the Proposed YOLO-World Model: (a) C3k2 Layer Flowchart, (b) Modified YOLO Backbone**

### 3.3. Experimental evaluation metrics

This paper uses precision (P), recall (R), F1 score, average precision (AP), and mean average precision (mAP) as the primary metrics for evaluating and comparing the experiments. Detailed definitions and formulas are given in equations (1) to (5), where several key terms are also defined. A true positive (TP) refers to a case where the model correctly identifies a positive instance. A false positive (FP) occurs when the model incorrectly classifies a negative instance as positive. A false negative (FN) is when the model fails to recognize a positive instance and classifies it as negative. A true negative (TN) is when the model correctly identifies a negative instance as negative.

Precision (P) is defined as the ratio of true positives to all instances classified as positive by the model. For example, in the context of identifying trucks in an image, it measures the proportion of predicted trucks that are actually trucks. Recall (R) is the ratio of true positives to all actual positive instances, indicating the proportion of real trucks that the model successfully detects. The F1 score is the harmonic mean of precision and recall. Because there is often a trade-off between precision and recall, the F1 score is particularly useful for evaluating model performance, especially when dealing with unbalanced data sets.

$$P = \frac{TP}{TP + FP} \tag{1}$$

$$R = \frac{TP}{TP + FN} \tag{2}$$

$$F1\,score = \frac{2 \times Precision \times Recall}{Precision + Recall} \tag{3}$$

The precision-recall curve is used to accurately evaluate the performance of metrics that have a trade-off relationship. Average Precision (AP) quantifies the performance of an object detection algorithm with a single value by calculating the area under the precision-recall curve. A higher AP value indicates better model accuracy. In this paper, the mean average precision (mAP), obtained by averaging the AP values across all classes, is used as the final metric for model evaluation.

$$AP = \int_0^1 P(r)\,dr \tag{4}$$

$$mAP = \frac{1}{N}\sum_{i=1}^{N} AP_i \tag{5}$$

## 4. RESULTS AND ANALYSIS

### 4.1. Experimental environment and parameter settings

The experimental environment was configured using Google Colab. Python (3.11.12) was used as the programming language, and PyTorch (2.6.0) served as the deep learning framework. CUDA version 12.4 was used and an NVIDIA Tesla T4 GPU was used. The system was equipped with 15 GB of RAM. The number of training epochs was set to 100, and the initial input image size was 640×640 pixels. In addition, the experiments were conducted using the default parameters (depth multiplier: 0.33, width multiplier: 0.50) and configuration settings of YOLOv8s. In addition, a stochastic gradient descent (SGD) with a learning rate of 0.01 and a momentum of 0.937 was used as the optimizer. The detailed specifications are summarized in Table 1.

**Tab. 1. Hardware and software parameters of the training system**

| Name | Parameters |
|---|---|
| Development environment | Google Colab |
| GPU | Nvidia, Tesla T4 |
| Installed RAM | 15 GB |
| CUDA Version | 12.4 |
| Programming language | Python 3.11.12 |
| Deep learning framework | PyTorch 2.6.0 |
| Optimizer | SGD lr = 0.01, momentum = 0.937 |

The dataset used in this experiment is based on the VisDrone dataset, one of the most reliable and widely used datasets for drone-based object detection. The VisDrone dataset is commonly used to evaluate the performance of object detection models using images and videos captured by drones. It consists of a total of 8,629 images, of which 6,471 were used for training, 548 for validation, and 1,610 for testing. The objects to be detected are categorized into ten classes, which are pedestrian, person, bicycle, car, van, truck, tricycle, awning-tricycle, bus, and engine. An example image from the dataset used in the experiment is shown in Figure 3.

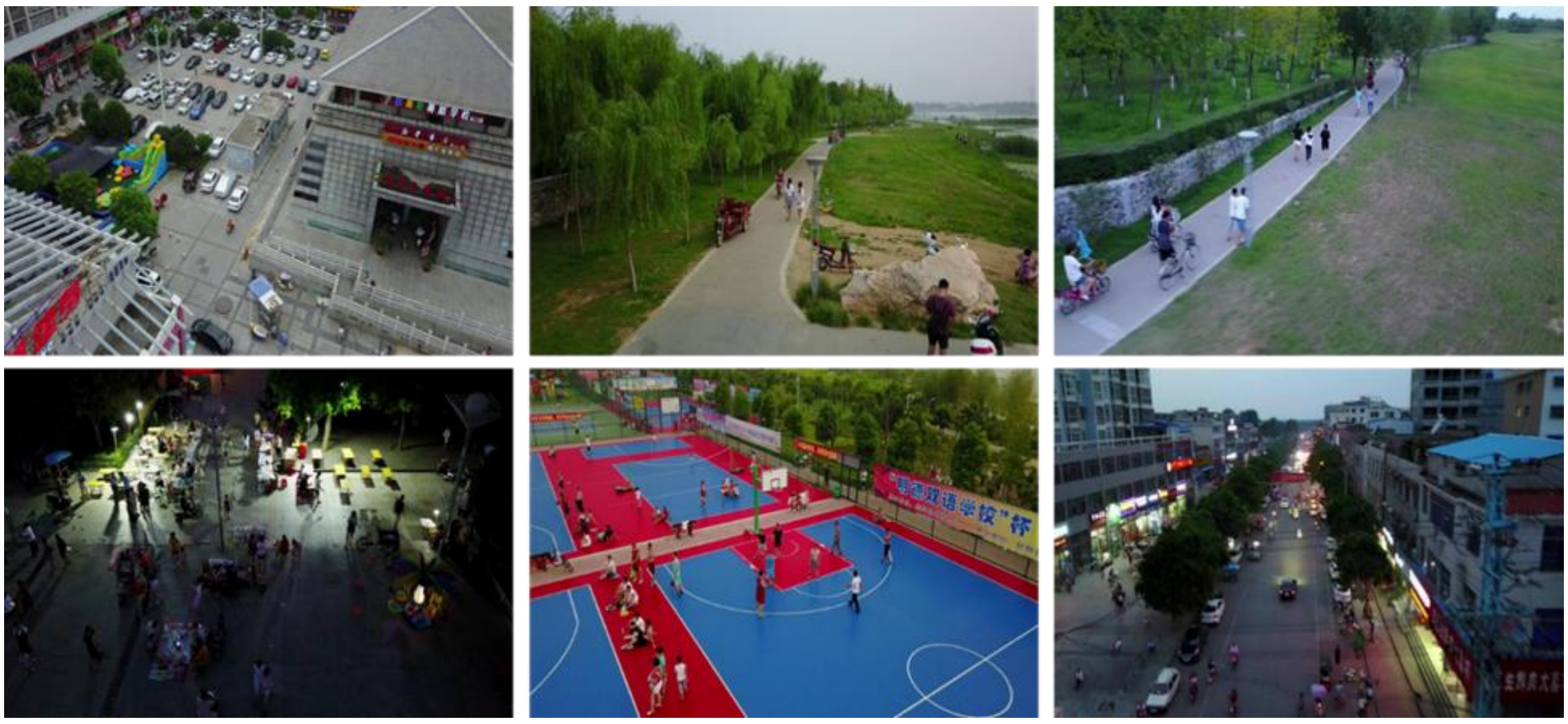

**Fig. 3. Sample Images from the VisDrone Dataset Used in the Experiment**

## 4.2. Experimental results

Figure 4 shows the types, number, and distribution of all classes included in the VisDrone dataset used in this experiment. Figure 4(a) shows the class names along with the corresponding number of instances, indicating that the dataset contains a sufficient number of samples for effective experimentation. Figure 4(b) shows the distribution of the object labels, where the x-axis represents the ratio of the label center to the image width, and the y-axis represents the ratio of the label center to the image height. This indicates that the labels are generally well distributed, with a noticeable concentration near the center of the image. Figure 4(c) shows the size of each class, where the x-axis represents the ratio of label width to image width, and the y-axis represents the ratio of label height to image height. Figure 5(a) shows the heatmap results of the C2f layer, while Figure 5(b) shows those of the C3k2 layer. These results visually demonstrate that the proposed C3k2 layer achieves improved performance in detecting small objects.

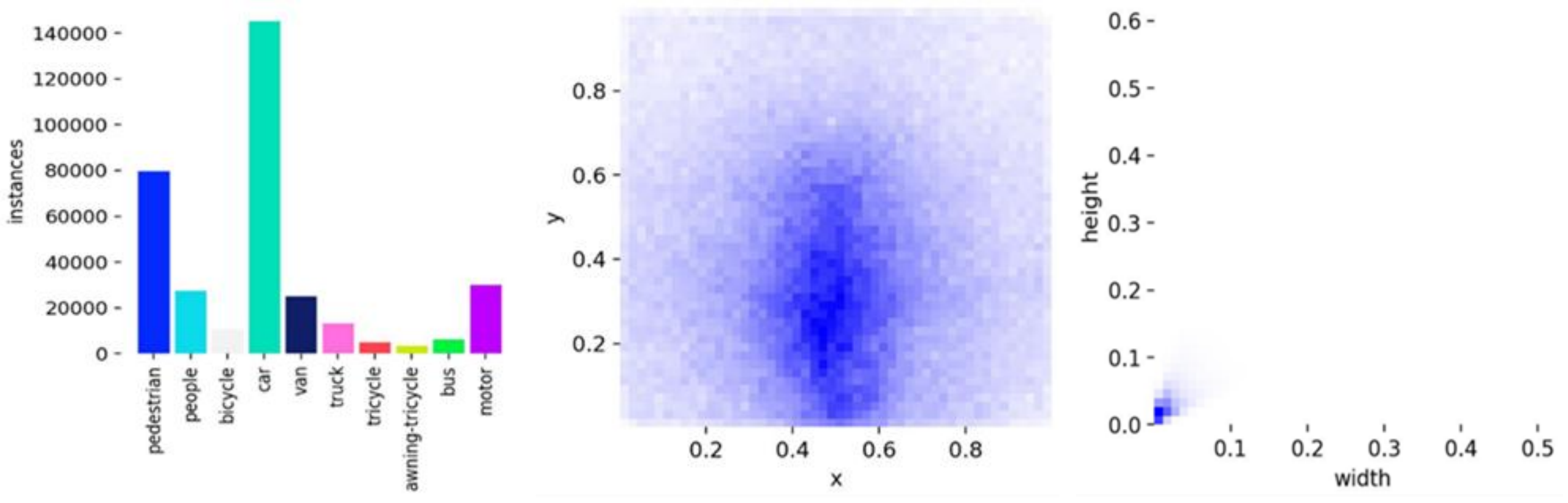

(a) (b) (c)

**Fig. 4. Number of Instances and Label Distribution for Each Class:**
**(a) Number of Instances, (b) Label Positions, (c) Label Sizes**

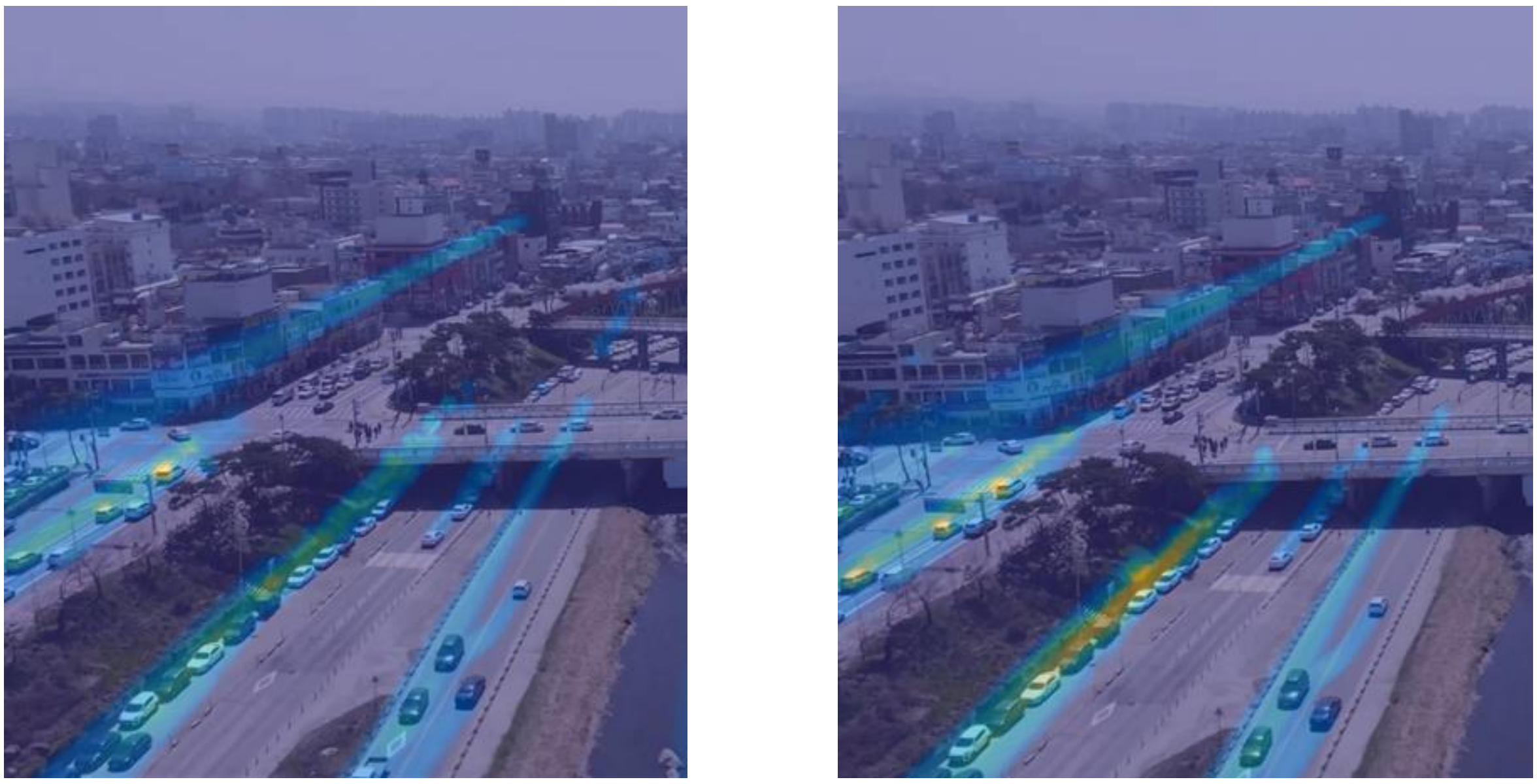

**(a)** **(b)**

**Fig. 5. Comparison of heatmaps for the C2f and C3k2 layers: (a) C2f layer, (b) C3k2 layer**

Figures 6(a) and 6(b) show the confusion matrix results of the YOLO-World model and the proposed model, respectively. As shown in Figure 6(b), the proposed model achieves high classification accuracy and prediction performance in all classes, especially with 9,552 correct predictions for the car class and 2,263 for the pedestrian class. These results indicate that the proposed model not only preserves critical information effectively, but also demonstrates excellent discriminative ability, especially for small objects.

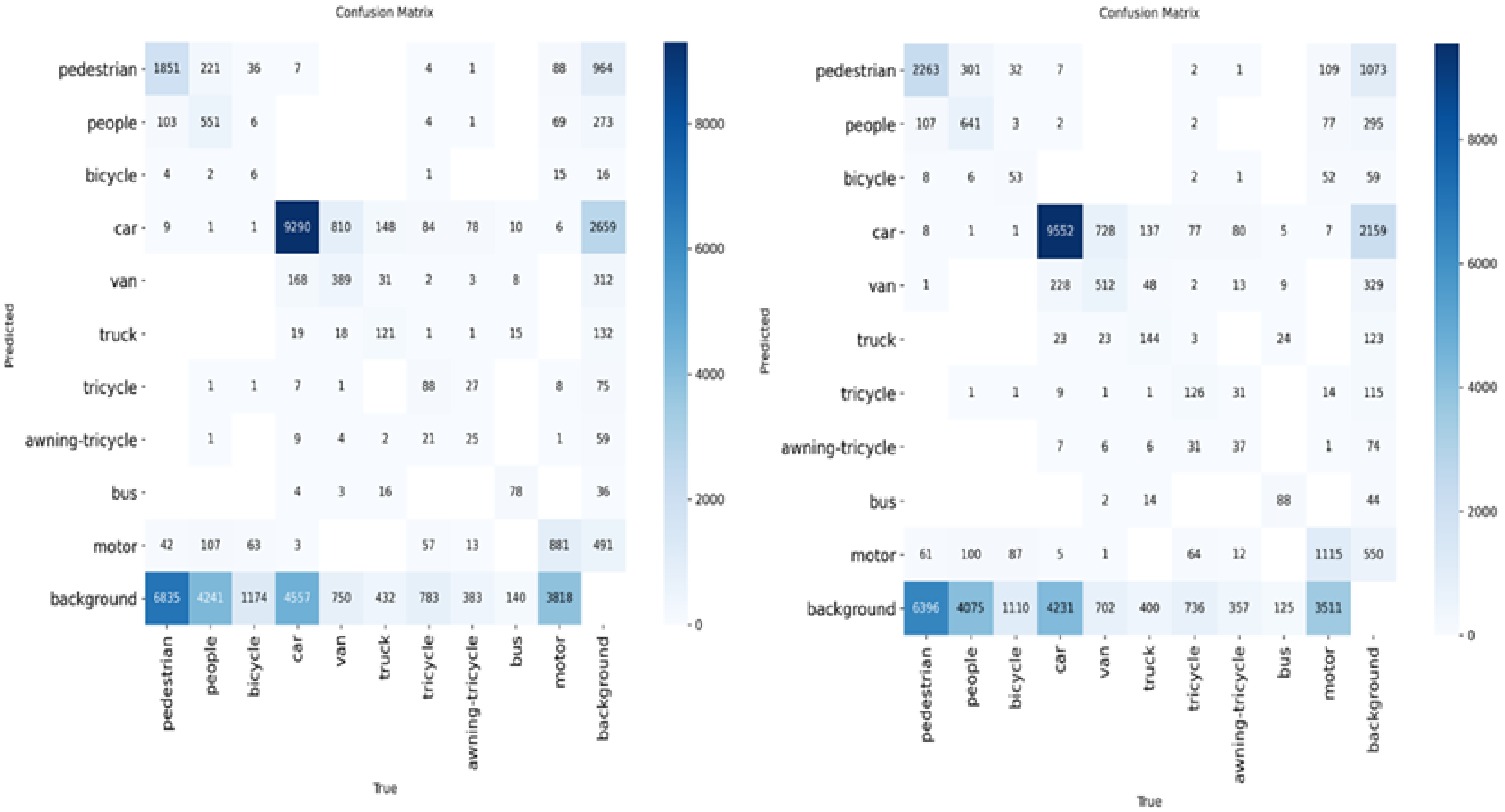

**(a)** **(b)**

**Fig. 6. Comparison of the confusion matrices of the YOLO-World model and the proposed model: (a) YOLO-World, (b) Proposed model**

The experiments were conducted independently using both the original YOLO-World model and the modified model, which incorporates a modified backbone network. Specifically, this variant replaces the C2f layers in the original YOLO-World backbone with C3k2 layers for comparative analysis. Evaluation metrics such as precision, recall, F1 score, and mAP@0.5 were then calculated to assess the accuracy of the models. The experimental results show that, compared to the original YOLO-World model, precision increased from 40.6% to 41.6%, an improvement of 1.0%. Recall increased from 30.8% to 31.0%, a gain of 0.2%. The F1 score improved from 35.0% to 35.5%, an increase of 0.5%. In addition, mAP@0.5 increased from 30.4% to 30.7%, an improvement of 0.3%. Reductions in FLOPs and the total number of parameters were also observed, indicating improved efficiency and a more lightweight design. Further, performance comparison experiments were conducted with state-of-the-art (SOTA) models in object detection and text-guided tasks, including YOLOv9, YOLOv10, YOLOv11, and the zero-shot detection YOLO model (Xie & Zheng, 2022).

**Tab. 2. Comparison of experimental results with various object detection models**

| Method | Vision Backbone | Text Model | Precision (%) | Recall (%) | F1 score (%) | GFLOPs | Params (M) | mAP@0.5 (%) |
|---|---|---|---|---|---|---|---|---|
| YOLOv9 | YOLOv9 | - | 42.3 | 30.2 | 35.2 | 7.9 | 2.0 | 30.2 |
| YOLOv10 | YOLOv10 | - | 40.4 | 30.3 | 34.6 | 8.2 | 2.6 | 29.7 |
| YOLOv11 | YOLOv11 | - | 40.2 | 30.6 | 34.7 | 6.3 | 2.5 | 29.8 |
| Zero-shot Detection YOLO | YOLOv5 | CLIP | 39.3 | 29.9 | 33.9 | 18.5 | 7.3 | 29.1 |
| YOLO-World | YOLOv8 | CLIP | 40.6 | 30.8 | 35.0 | 15.7 | 4.0 | 30.4 |
| Proposed Model | Proposed YOLOv8 | CLIP | 41.6 | 31.0 | 35.5 | 15.2 | 3.8 | 30.7 |

**Tab. 3. Comparison of experimental results for each class in the proposed model**

| Class | Instances | Precision (%) | Recall (%) | F1 score (%) | mAP@0.5 (%) |
|---|---|---|---|---|---|
| All | 38759 | 41.6 | 31.0 | 35.5 | 30.7 |
| Pedestrian | 8844 | 42.6 | 31.2 | 36.0 | 31.3 |
| People | 5125 | 48.9 | 21.0 | 29.3 | 25.5 |
| Bicycle | 1287 | 21.1 | 9.5 | 13.1 | 6.8 |
| Car | 14064 | 63.8 | 73.6 | 68.3 | 74.0 |
| Van | 1975 | 43.6 | 36.9 | 39.9 | 36.2 |
| Truck | 750 | 38.9 | 23.1 | 28.9 | 23.9 |
| Tricycle | 1045 | 35.9 | 21.8 | 27.1 | 19.5 |
| Awning-Tricycle | 532 | 25.0 | 13.2 | 17.2 | 11.0 |
| Bus | 251 | 51.3 | 45.0 | 47.9 | 44.2 |
| Motor | 4486 | 45.2 | 35.2 | 39.5 | 34.1 |

The results are summarized in Table 2. Table 3 presents the experimental results for each class evaluated using the modified model. Relatively low mAP@0.5 scores were observed for the bicycle and awning-tricycle classes, suggesting that these objects are more difficult to distinguish and are underrepresented in the dataset. In other words, their lower mAP@0.5 scores indicate that these objects pose a greater challenge for recognition. Additional experiments were conducted by inputting sentences into each model. Four prompts were provided, corresponding to the truck, pedestrian, car, and engine classes. The comparative results of the text-guided object recognition tasks are detailed in Table 4.

**Tab. 4. Comparison of object detection results with text input**

| **Methods** | **Command 1: Please find where the truck is** | **Command 2: I'd like to know where the pedestrian is** | **Command 3: I'm wondering where the car is** | **Command 4: Show me where the motor is** |
|---|---|---|---|---|
| YOLO-World | 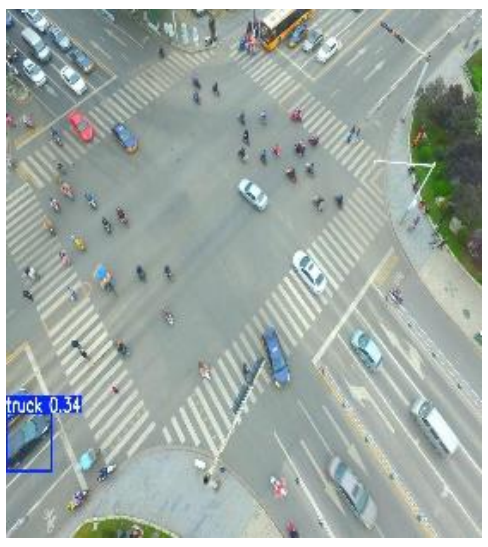 | 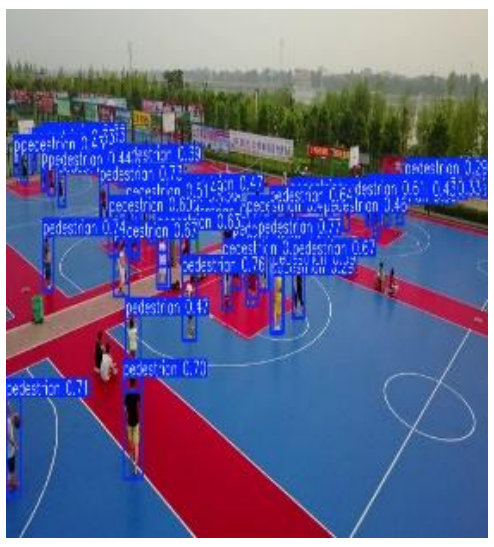 | 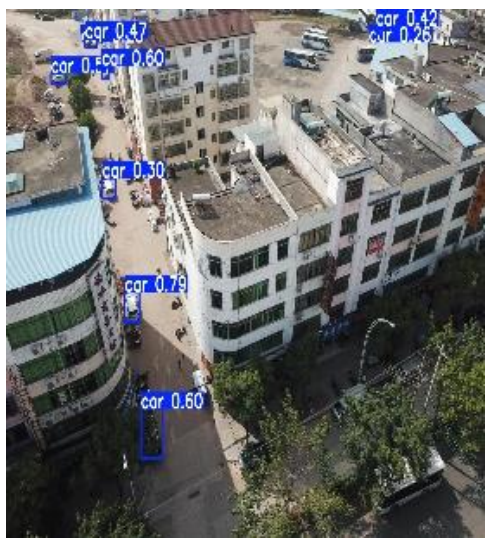 | 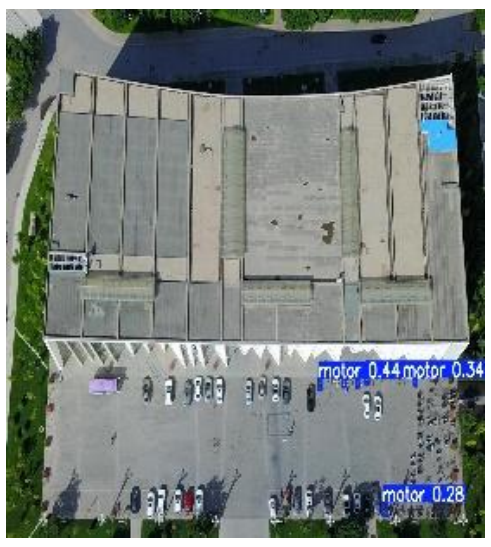 |
| Proposed Model | 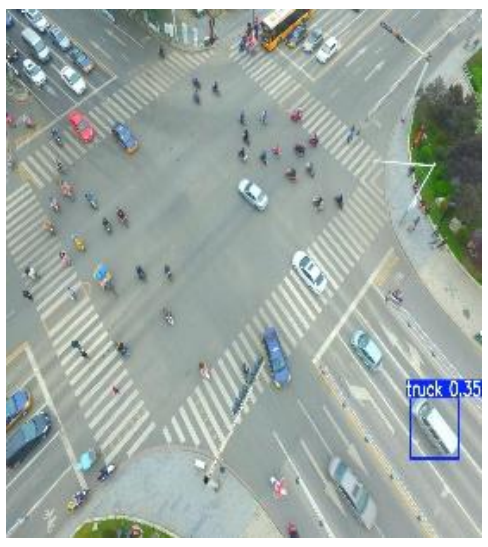 | 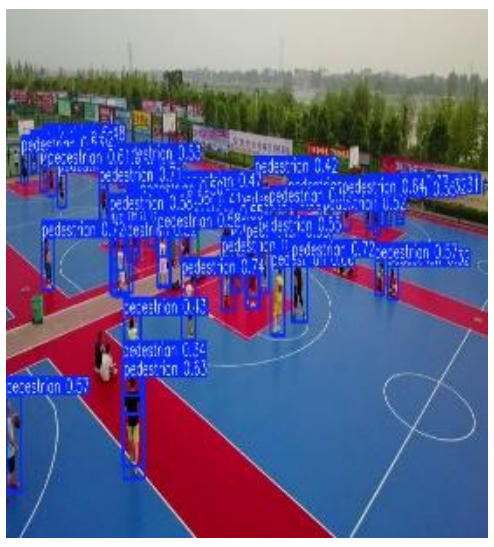 | 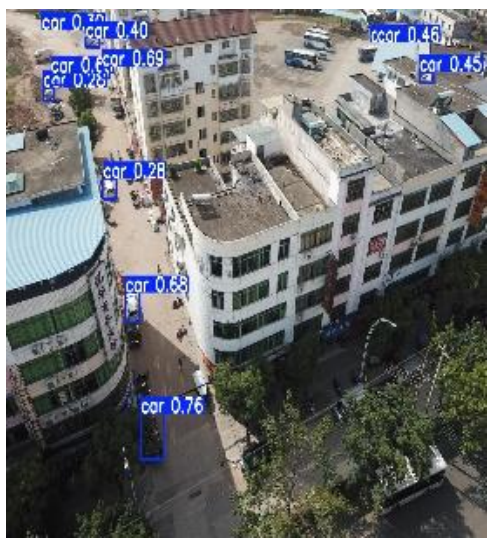 | 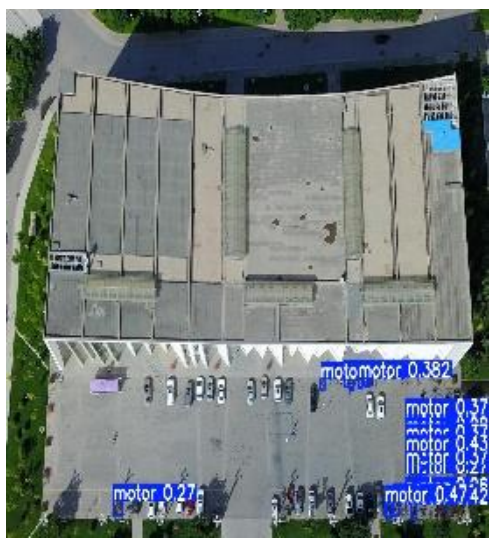 |

In addition, Figure7(a) on the left shows the precision-recall curves for each class of the proposed model. The awning-tricycle class had the lowest value of 0.11, while the car class had the highest value of 0.74. The overall recall value for all classes was 0.307. Furthermore, Figure7(b) shows the mAP@0.5 and loss values per epoch during training, confirming that the training of the model proceeded normally.

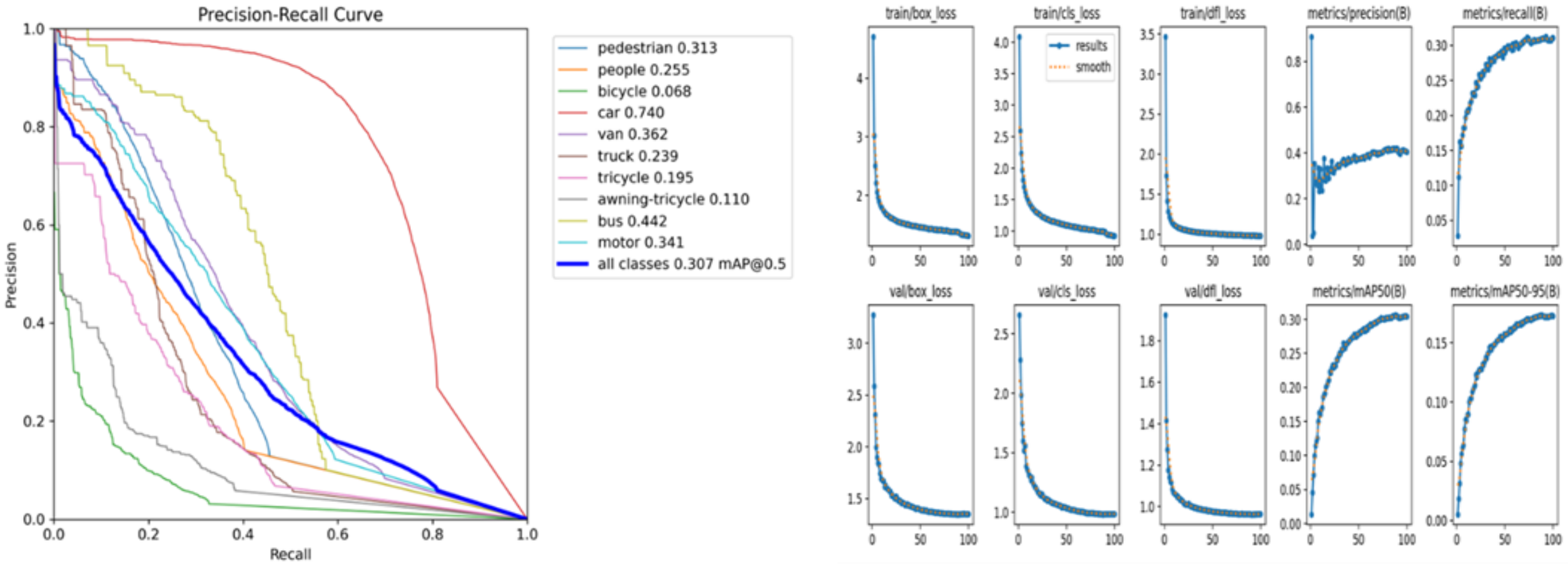

**(a)** **(b)**

**Fig. 7. Experimental results of the proposed model:**
**(a) Precision-recall curve, (b) Changes in key metrics across training epochs**

## 5. CONCLUSIONS AND LIMITATIONS

This paper deals with a text-guided object detection model designed to efficiently detect small objects using drone images and videos. The proposed improvement to YOLO-World is to extend the YOLOv8 backbone network previously used for feature extraction. Experiments have been conducted to validate this approach. The dataset used for the experiments was based on the VisDrone dataset, which includes a total of ten classes such as pedestrian, human, bicycle, car, van, truck, tricycle, awning-tricycle, bus, and engine. The proposed backbone network replaces the previously used C2f layers with C3k2 layers. Accordingly, the study aimed to improve the accuracy and model lightweight efficiency by developing a new backbone network.

The experimental results showed that the proposed method outperformed the original approach in terms of detection accuracy, achieving a mAP@0.5 of 30.7%, which represents a 0.3% improvement over the original model. In addition, improvements in FLOPs and number of parameters were observed. These results show that the proposed model is both lighter and more efficient than the original.

This study is expected to provide effective engineering solutions for various industrial and research applications using drones in the future. Although the proposed model shows excellent performance in detecting small objects, its performance may still degrade in certain challenging scenarios. For example, in cases of severe occlusion, the model may fail to fully capture the discriminative features of partially visible objects. Similarly, environmental variations such as weather conditions can reduce the efficiency of feature extraction. In high object density environments, overlapping targets can lead to missed or incorrect detections due to spatial ambiguity. To overcome these limitations, future work will focus on improving generalization under adverse conditions by integrating attention mechanisms into the backbone or head network architecture of the model.

## Funding

*This research received no external funding.*

## Data Availability Statement

*All the data used in the experiments was based on the VisDrone: 2019 dataset.*

## Conflicts of Interest

*The author declares no conflict of interest.*